\documentclass[letterpaper,10pt,conference]{ieeeconf}
\IEEEoverridecommandlockouts
\usepackage{amsmath}
\usepackage{amssymb}
\usepackage{caption}
\usepackage{color}
\usepackage{graphicx}
\usepackage{subcaption}

\title{\LARGE \bf
    Collision avoidance from monocular vision \\
    trained with novel view synthesis
}

\author{%
    Valentin Tordjman{-}{-}Levavasseur$^{1}$ and Stéphane Caron$^{1}$
    \thanks{$^{1}$ The authors are with Inria and the Computer Science Department of ENS (DI ENS), PSL Research University, Paris, France. Corresponding author: {\tt\small valentin.tordjman-levavasseur@inria.fr}}%
}

\begin{document}

\maketitle
\thispagestyle{empty}
\pagestyle{empty}


\begin{abstract}
Collision avoidance can be checked in explicit environment models such as elevation maps or occupancy grids, yet integrating such models with a locomotion policy requires accurate state estimation. In this work, we consider the question of collision avoidance from an implicit environment model. We use monocular RGB images as inputs and train a collision-avoidance policy from photorealistic images generated by 2D Gaussian splatting. We evaluate the resulting pipeline in real-world experiments under velocity commands that bring the robot on an intercept course with obstacles. Our results suggest that RGB images can be enough to make collision-avoidance decisions, both in the room where training data was collected and in out-of-distribution environments.
\end{abstract}


\section{Introduction}

Understanding and interacting with its environment is critical for a robot performing locomotion tasks. Notably, in order not to fall, it should identify and avoid colliding with obstacles on its way. This can be tackled using explicit representations such as occupancy grids or elevation mapping. Yet, these approaches can struggle when dealing with unstructured elements like leaves or grass in real-world deployments~\cite{miki2022science}. Additionally, they tend to require more expensive hardware such as LiDARs, and introduce computational overhead such as point-cloud processing, making them challenging for real-time control.

On the other hand, recent work on implicit scene representations addresses some of these challenges by learning higher dimensional encodings of the environment without requiring explicit reconstruction. These approaches leverage learned features from raw sensory data. However, generating realistic RGB image data in simulation remains a major challenge, as domain discrepancies between synthetic and real-world images introduce a substantial sim-to-real gap. As such, depth image as been used more extensively to train an implicit scene representation in simulation, as depth data tends to suffer from a smaller sim-to-real gap compared to unrealistic RGB textures. However, this strategy either relies on expensive sensors, or suffer from the poor measurements of cheaper sensors.

To incorporate RGB image data, \cite{loquercio2022learningvisuallocomotioncrossmodal} train first a blind quadruped locomotion policy in simulator, and train the policy with vision using image data collected on the real robot. While successfully exhibiting a better anticipation and smoother locomotion using visual data, it would be more convenient to generate to realistic RGB image data directly in simulator. Another recent approach \cite{byravan2023nerf2real} uses neural radiance fields (NeRF) to have access to photorealistic RGB images in simulator, at the cost of a longer training time, as NeRF are known to be computationally intensive.

In this paper, we present a vision-guided velocity correction pipeline for collision avoidance, depicted in Fig.~\ref{pipeline}. To do so, we leverage the ability to render images in real time rendering speeds of Gaussian splatting techniques to train efficiently a high level locomotion policy, that will correct the joystick input of a (reckless) user, so that the robot avoids obstacles. The policy observation consists only on the joystick user input, the robot velocity, and image features, and uses no history or recurrent policy. The vision part consists of an autoencoder-like architecture which is trained separately on the task of monocular depth prediction on synthetic image data. These choices enable us to deploy the policy on an Upkie robot which has no dedicated compute unit. The high-level locomotion policy corrects joystick inputs of the user to control a separate locomotion controller.

Our main contributions are 1) a visual learning pipeline for collision avoidance based on novel view synthesis, 2) showing that a collision-avoidance policy can be successfully trained using only monocular RGB images as inputs, and 3) evaluating on a real robot the repeatability and ability of this approach to generalize beyond training data.

\begin{figure}[t]
    \centering
    \includegraphics[width=0.99\linewidth]{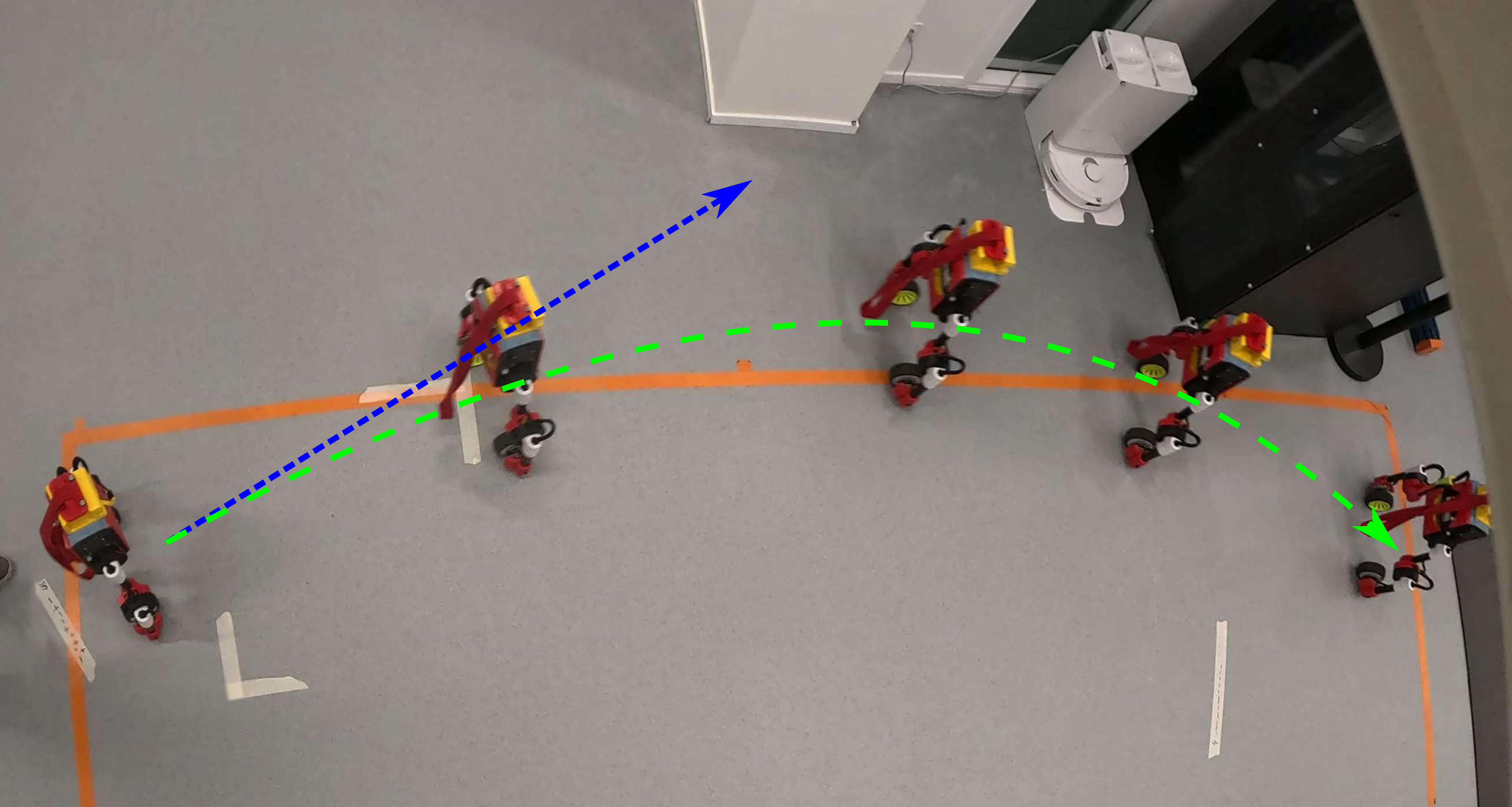}
    \caption{
        Effect of the vision-based collision-avoidance policy when the commanded velocity prompts the robot to collide with a wall. Blue: joystick user input, kept stationary at full forward throttle. Green: trajectory actually followed by the robot after compensation by the policy.
    }
    \label{vignette}
\end{figure}
\begin{figure*}[ht]
    \centering
    \includegraphics[width = \linewidth]{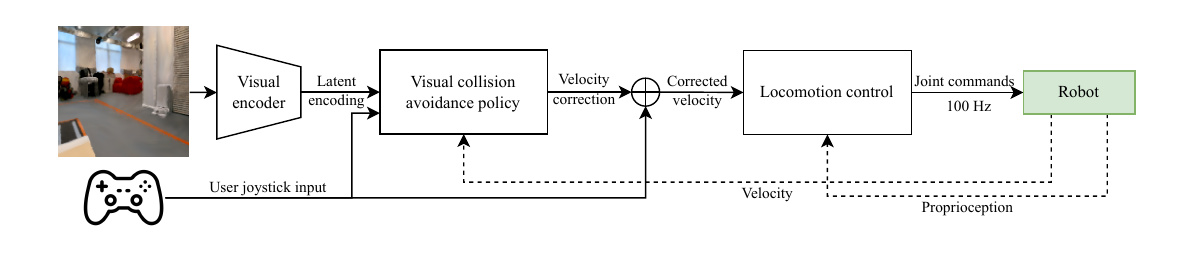}
    \caption{Monocular obstacle avoidance pipeline from the perception to the joint commands.}
    \label{pipeline}
\end{figure*}
\section{Related works}

The computer graphics literature commonly treats scenes while the robotics literature considers environments. A scene denotes the geometric state of the robot's environment at a given time, whereas an environment may more generally respond kinematically (velocities, accelerations, ...) or dynamically (inertias, friction, ...).

\subsection{Explicit scene representations}

Several models have been proposed to map sensory measurements to scene representations that are actionable for locomotion.

\subsubsection{Elevation mapping}

For ground locomotion in structured environments, an \emph{elevation map} decomposes the horizontal plane into cells and stores the ground altitude corresponding to each one~\cite{herbert1989}. This representation informs the robot as to whether it can step onto nearby location, or should steer away from areas such as walls or gaps. It is not in itself well-suited to unstructured or non-rigid terrains~\cite[Fig. 2]{miki2022science}, a characteristic that has prompted follow-up post-processing methods such as recurrent encoding trained from simulated data~\cite{miki2022science}. Elevation maps are updated at a relatively low 10-20 Hz frequency~\cite{fankhauser2018probabilistic} and are sensitive to state estimation errors~\cite{miki2022iros}. While we don't discard their applicability, in this work we rather explore a different sensor (RGB camera) and an end-to-end differentiable control pipeline that does not rely on intermediate state estimation.

\subsubsection{Occupancy grids}

Occupancy grids are another volumetric approach that decompose 3D Euclidean space into voxels and store in each one whether it is free space or belongs to an obstacle~\cite{hornung2013octomap}. This approach was combined with nonlinear model predictive control in~\cite{jacquet2024n} using a occupancy network~\cite{mescheder2019occupancy}, \emph{i.e.}~a neural-network approximation of an egocentric occupancy grid. The network receives as input a depth image $\mathbf{i}$ and produces, by means of a variational autoencoder (VAE), a differentiable function $\mathbb{E}^3 \to \mathbb{R}, \mathbf{p} \mapsto g_{\theta}(\mathbf{p}; \mathbf{i})$ that can be used as constraint in a nonlinear optimal control problem. In this work, we train a variational encoder decoder, but on RGB images generated by Gaussian splatting rather than from depth images generated from environment meshes. Our policy is a neural network, rather than a nonlinear optimization, that operates directly on a low-dimensional latent state computed by the vision encoder. This means multiple images can, for instance, be included in the observation history.

\subsubsection{Signed distance fields}

Rather than occupancy, the discretized environment model can store signed distances to the closest obstacles, resulting in a \emph{Euclidean signed distance field}. Finite differences from neighboring voxels can then approximate the gradient of the underlying signed distance function for inclusion in a nonlinear model predictive controller. This approach is followed in~\cite{gaertner2021collision} to add collision avoidance to the receding horizon objective. One downside with signed distance fields and occupancy grids is that the time complexity of data-structure updates is cubic in the voxel size, requiring a trade-off between environment-model accuracy and update frequency. The approach we follow in this work prompts different decisions: rather than targeting accurate reconstruction of an environment model that is then used for control, we directly process input images into actionable control inputs.

A shared attribute of locomotion pipelines that rely on explicit scene models is the inclusion of state estimation, either as the floating-base pose~\cite{bloesch2015robust, benallegue2017tilt} or more generally an element of the robot configuration space~\cite{hartley2020contact}. Even an ideal reconstruction with ten-fold the performance of the current state of the art (\emph{e.g.} centimetric precision and 500~Hz frame rate) would not warrant more reliable locomotion unless state estimation were to improve jointly. The explicit approach thus puts a high-accuracy requirement on low-dimensional state coordinates. In contrast, designs with implicit scene representations, such as the one we propose thereafter, can rely on higher-dimensional latent states with higher redundancy and less stringent requirements on individual coordinates.

\subsection{Implicit scene representations}

In recent works, depth images have been used more extensively than RGB images as sensory inputs. One motivation is that rendering depth images is less computationally and implementation-intensive than generating photorealistic RGB images. Distance measurement sensors such as lidars and RGB-D cameras can be mounted on real-robots, although the latter are not well-suited to working outdoors and the former are overall more expensive.

\subsubsection{Distillation of privileged state information}
 
Teacher-student distillation is a generic method where a first policy is trained in simulation with access to privileged information (exact distances to contacts, surface friction coefficients, etc.) before a second policy is trained to act similarly to the first one yet with only access to observable quantities. Distillation was already proposed to train blind locomotion policies in~\cite{lee2020learning}, with the design reproduced in RMA~\cite{kumar2021rma}, where privileged information is encoded to a latent state in the teacher architecture, and the student policy is trained to infer the same latent states from sensor observations only. This approach was extended in~\cite{agarwal2023legged} by adding depth images to the sensory inputs used to infer latent states. Alternatively to having the teacher policy learn its own latent states, SoloParkour~\cite{chane2024soloparkour} learns the student policy by offline reinforcement learning with a replay buffer populated in part by teacher actions annotated with depth images.

These approaches learn locomotion policies end-to-end, where the main rewarded behavior is to track a prescribed body velocity. They don't suppose that the user or software deciding that velocity may, willfully or not, take the robot to a fall-prone state. This raises the question of adversarial commands~\cite{shi2024rss}. In this work, we focus on collision avoidance, \emph{e.g.}~when the commanded velocity takes the robot on an intercept course to a wall or a standing obstacle. We design a visual learning pipeline to train a policy that compensates collision-prone commands.

\subsection{Novel view synthesis}

Novel view synthesis is a computer graphics task where, given a set of input images, we generate images of the same scene from different viewpoints. Methods for this have evolved from photogrammetry to deep learning. Neural Radiance Fields (NeRF)~\cite{mildenhall2021nerf} are a technique of the latter kind where the scene is represented as a continuous 3D radiance field using a neural network. Yet, this technique relies on ray tracing to render images, which is computationally expensive and typically runs at less than 1~Hz~\cite{garbin2021fastnerfhighfidelityneuralrendering}.

Moving forward, 3D Gaussian splatting (3DGS) introduced an explicit scene representation through 3D Gaussians. Training is done using gradient descent to fit 3D Gaussian distributions to the scene, enabling real-time rendering (faster than $100$~Hz) thanks to the closed-form rasterization of Gaussians. 3DGS may fail to accurately represent surfaces as Gaussians are prone to multi-view inconsistency. To circumvent this, 2D Gaussian splatting~\cite{Huang_2024} proposed fitting 2D Gaussians (surfels) to the 3D scene, introducing normal consistency and depth distortion regularization to achieve geometrically accurate radiance fields.


In~\cite{byravan2023nerf2real}, NeRFs are applied to generate realistic RGB images coupled with a physics simulation~\cite{todorov2012mujoco}. The resulting pipeline is used to train end-to-end navigation and ball-pushing policies that act on both visual and proprioceptive inputs. In this work, we apply 2DGS rather than NeRFs for faster rendering, as it directly translates into faster training times. Rather than an end-to-end recurrent policy, we first train a visual encoder, then a collision-avoidance policy whose outputs are forwarded to a model-predictive locomotion controller.

\section{Scene representation and rendering}

\subsection{Dataset collection}

Since our pipeline consists of various tools to achieve novel view synthesis in the simulator, the quality of the data used is critical. Moreover, as COLMAP tends to be quite sensitive, poor-quality images could lead to incorrect pose estimation. In our experience, a key takeaway for good-quality Gaussian splats is to keep an eye out for motion blur. To achieve this, we set the shutter speed of our camera to at least $1/125$~s, which implies correspondingly large ISO and aperture settings to avoid underexposure in our indoor scene. We used a GoPro camera set to auto-focus with a wide lens of 16~mm.

Camera motion during capture is also important. Standard photogrammetry recommendations apply: we need to induce as much parallax as possible by hovering around the objects we want to capture. We typically capture a 5-minute video for a single room.

Due to video file compression, the resulting file has keyframes (actual pictures) and interframes (interpolations of keyframes). We use FFMPEG to extract the key frames, as they are of better quality and tend to have less motion blur.  We capture in 4k 60 fps to have as many keyframes as possible. A manual check is performed to remove any remaining blurred images. This process typically results in around 500 images.

Once our images cover the scene extensively, we use COLMAP~\cite{schoenberger2016sfm} to estimate camera poses, the intrinsics of the camera, and a sparse point cloud of the scene. Since the COLMAP reference frame is arbitrary, lacks scale, and does not align the $z$-axis with gravity, we take at least three pictures from known camera positions. In NeRF2Real~\cite{byravan2023nerf2real}, the $z$-axis is aligned with gravity using a least square to find the ground plane, and scaled manually in Blender. The alternative we chose was to use corners of the room or tables to easily define relative positions of three images. This allows COLMAP to perform geo-registration, i.e estimate the transform between the arbitrary reference frame and the target reference frame. This produces a gravity-aligned $z$-axis and a correctly scaled reference frame. 

\subsection{Integration with a physics simulator}

\begin{figure}[t]%
    \centering
    \subfloat[\centering Raw mesh export]{{\includegraphics[width=0.42\linewidth]{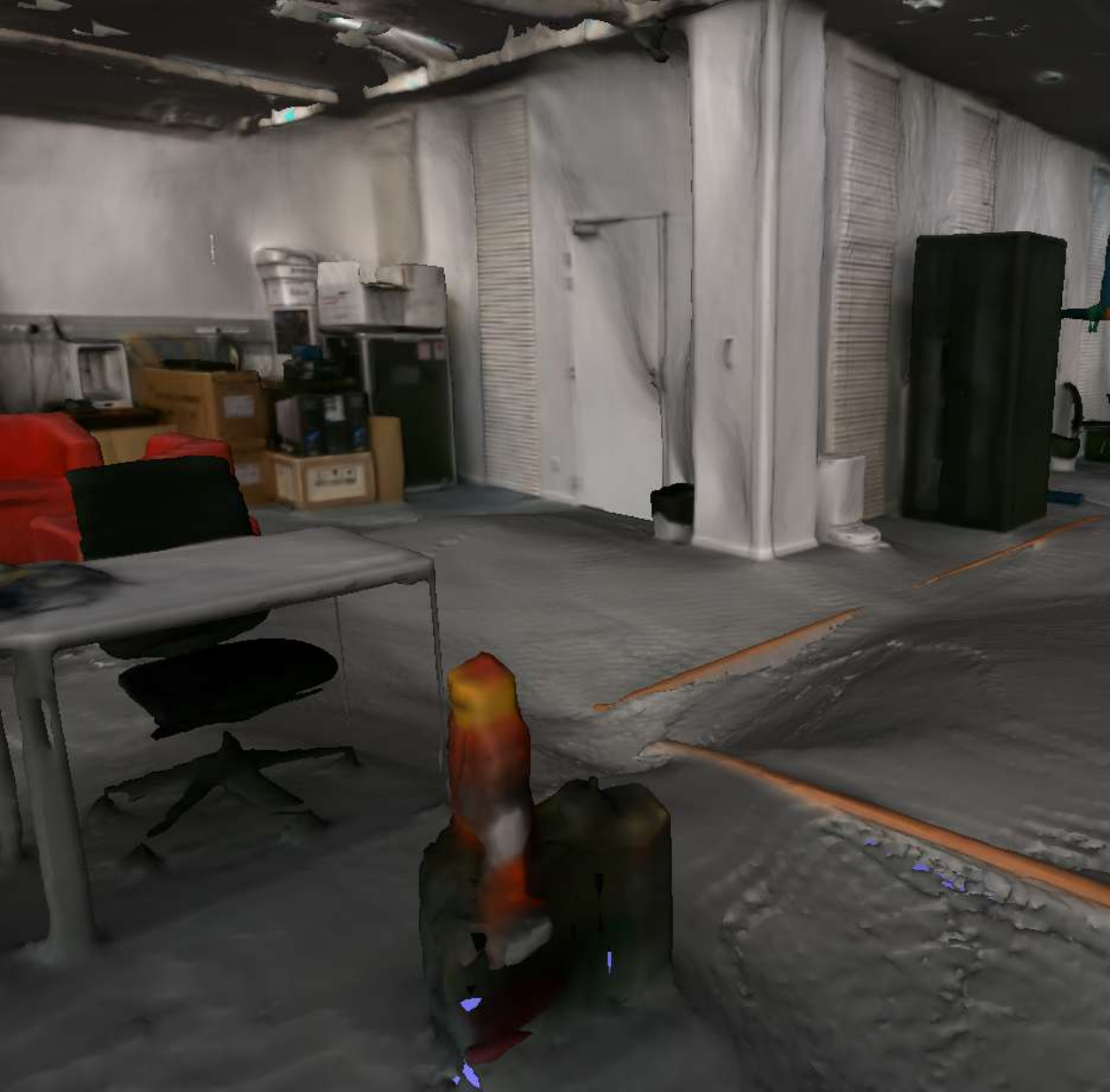} }}%
    \qquad
    \subfloat[\centering Collision mesh]{{\includegraphics[width=0.42\linewidth]{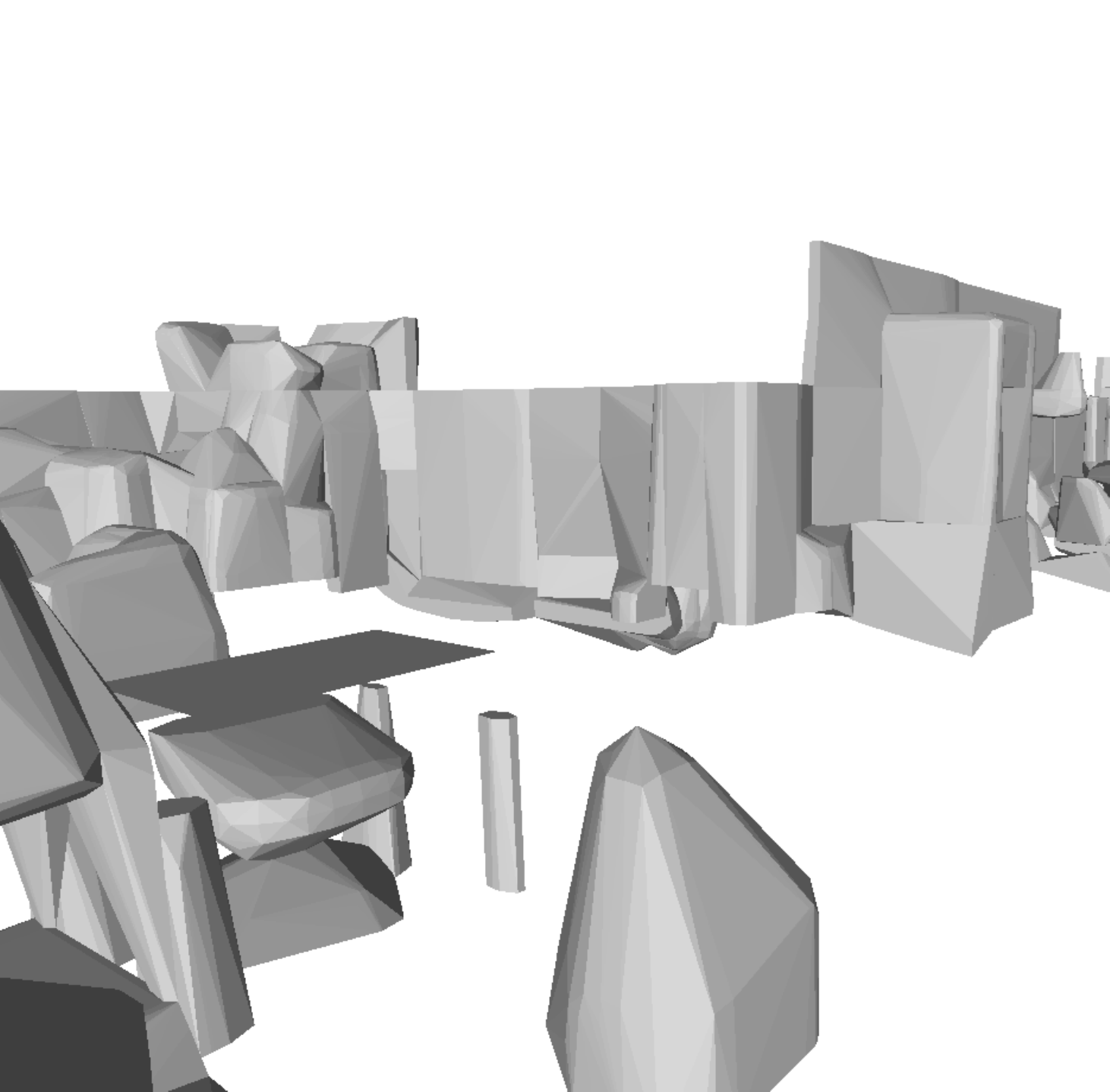} }}%
    \caption{Comparison between the raw mesh export and the mesh after being processed by CoACD, and manually cleaned up.}%
    \label{comparaison mesh}%
\end{figure}

With the dataset collected, we train the 2DGS model, and extract the 3D mesh of the scene using the same mesh extraction pipeline as 2DGS~\cite{Huang_2024}. We then load the scene mesh and a model of the Upkie wheeled biped in Pinocchio~\cite{pinocchioweb}. The robot camera's field of view is then passed to the rasterizer, generating first-person robot camera views. To detect collisions with the environment, we decompose the scene mesh into convex sub-parts using collision-aware convex decomposition (CoACD)~\cite{coacd}.
Given the poor mesh quality on textureless surfaces, especially the floor, we manually replaced the convex subparts for the floor and tables with planes in Blender to ensure better simulation accuracy. We compare the post-processed mesh with the raw extraction in Fig.~\ref{comparaison mesh}. With the mesh decomposed and loaded as a collection of convex subparts, we use the open-source Coal library~\cite{coal} to compute distances between the robot and the environment.

\section{Training collision-avoidance policies}

\subsection{Navigation environment}

Our approach is to train a neural network to apply corrections to the user joystick inputs to avoid obstacles. This agent will run at 10Hz, while the locomotion controller will run at 100~Hz. To do so, we built a pure navigation environment, which observation space is the joystick user input, the current velocity, and the visual features of a visual encoder. The action space is the joystick correction. The corrected joystick is the sum of the user input and the correction.

The agent task in the environment is to apply as little joystick correction as possible, while surviving. For example, going forward: 
\begin{itemize}
    \item If no obstacle is to be seen, no correction is needed
    \item If an obstacle is closing in, but the robot sees at least one clear way of continuing to go forward, it must go toward the clear way of least deviation
    \item If the robot is in a dead end, it must either stop or turn around
\end{itemize}
We defined the reward as such:
\begin{equation}
R(s, a, s') =
\begin{cases}
1 - \| a \|_1 & \text{if $s'$ is collision-free} \\
-100 & \text{otherwise}
\end{cases}
\end{equation}
With $a$ being the action, a normalized velocity correction.
The environment terminates if the agent is less than 2 centimeters away from any obstacle. A larger margin could have been used for a more conservative policy, as we observe some risky behaviour where the agent tries to pass between narrow obstacles at evaluation. 

The agent is spawned is the scene randomly in the scene at least 50 centimeters away from the obstacles, and follows a simple dynamic made to mimic what the agent will have to deal with when controlling the locomotion controller afterwards. At each timestep, the agent updates its state \( \mathbf{x} = (x, y, \theta) \), where \(x, y\) denote planar coordinates and \(\theta\) represents the orientation. The velocity evolves subject to rate constraints \(|\dot{v}| \leq a_{max} \Delta t\), with \(a_{max}\) the maximum linear acceleration of the locomotion controller, and additive Gaussian noise. The velocities are also clipped to the maximum commanded velocities of the locomotion controller. Linear velocity is projected onto the world frame using the agent’s orientation, resulting in the kinematic update:
\begin{align}
    x_{t+1} &= x_t + v_x \cos(\theta_t) \Delta t \\
    y_{t+1} &= y_t + v_x \sin(\theta_t) \Delta t \\
    \theta_{t+1} &= \theta_t + v_\theta \Delta t
\end{align}
Additionally, to emulate the natural tilt of the inverted wheeled robot, the image pitch angle is perturbed by zero-mean Gaussian noise and constrained within \([- \frac{\pi}{4}, \frac{\pi}{4}]\) rad. We also apply data augmentation in contrast, brightness, hue, saturation, and noise.

\subsection{Navigation agent}

\begin{figure}[t]
    \centering
    \includegraphics[width=0.99\linewidth]{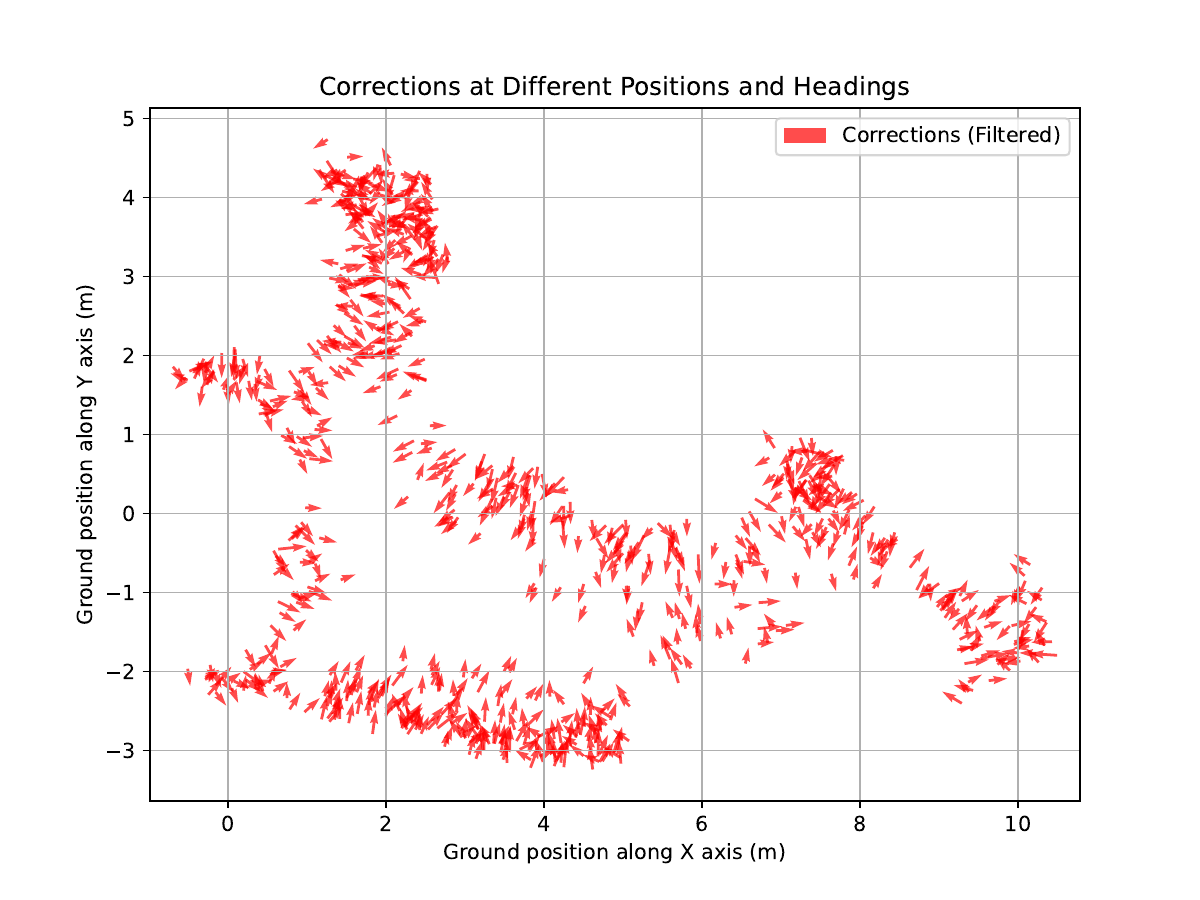}
    \caption{Most significant corrections applied by the collision-avoidance policy in the pure navigation environment when prompted to go fully forward. The most significant corrections are applied near the walls, away from them.}
    \label{fleches rouges}
\end{figure}

We train the navigation agent  using CrossQ~\cite{crossq}, implemented in Stable Baselines~3~\cite{sb3}, to improve sample efficiency while maintaining stable learning. CrossQ enhances off-policy reinforcement learning by leveraging Batch Normalization in the critic network and removing the need for target networks, allowing for efficient value estimation with a low update-to-data (UTD) ratio. Given that generating images makes each environment step computationally expensive compared to a physics engine step, CrossQ's ability to learn effectively from fewer interactions is particularly beneficial.
For the policy, we use a two layer multi-layer perceptron (MLP) of size 256 with batch norms, and for the critic a two layer MLP of size 1024 with batch norms. We train the policy on one environment for 500,000 steps, which takes 2.5 hours on an NVIDIA® GeForce RTX™ 4070 Ti SUPER.

At the end of the training, in Fig.~\ref{fleches rouges} we plot the most significant corrections given by the agent when running for 10,000 timesteps.

\subsection{Handling vision}

We decided to train the visual perception part separately. This makes it easier to debug, as we can check if the vision part successfully trains. To do so, we first collect image data using mock runs in the navigation environment. We collect 65,000 RGB and depth images of size $128 \times 128$ px, 60,000 of which will constitute the training set, and the rest will be the test set. This data collection step takes about 20 minutes on a consumer computer. Note that collecting the depth along with the RGB image is free as it is already rendered by the Gaussian splatting renderer.

\begin{figure}[t]
    \centering
    \includegraphics[width=0.99\linewidth]{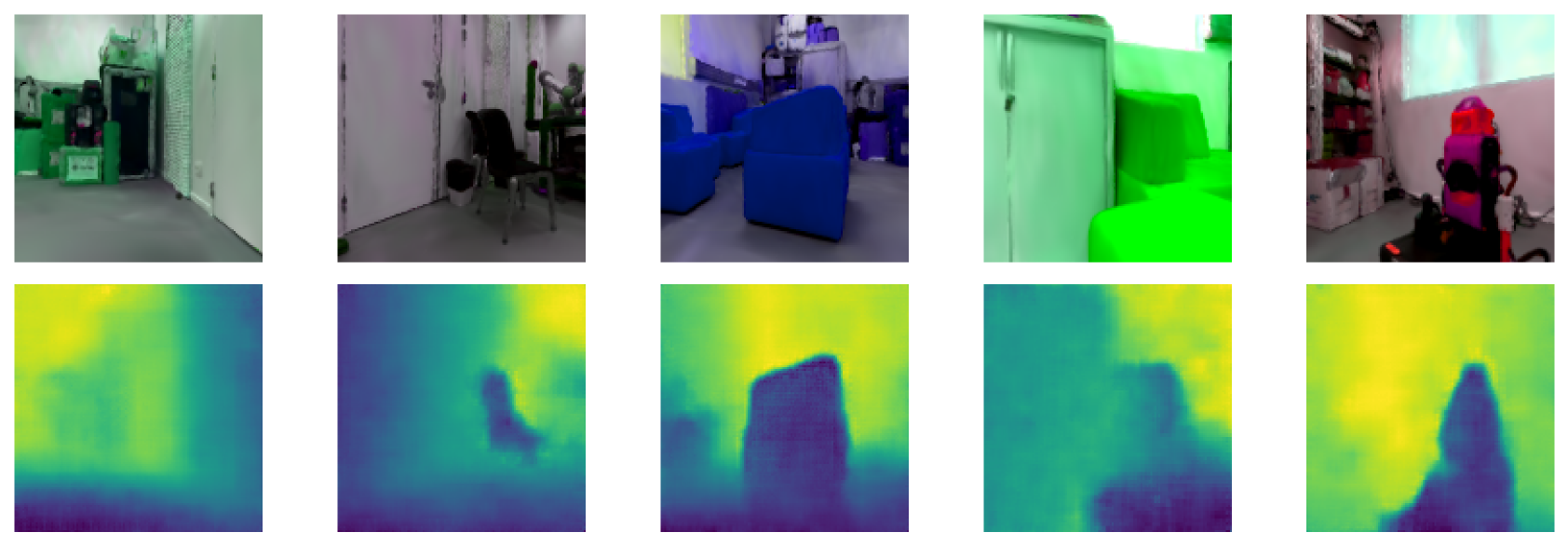}
    \caption{Examples of depth reconstruction. On the first line is the RGB image given to the encoder, on the second line the depth output of the decoder.}
    \label{depth reconstruction}
\end{figure}

We then train an autoencoder-like convolutional neural network (CNN) to learn the visual encoder to map the RGB image to the log of the depth image. Choosing the log space for depth values helps to balance the distribution of errors, as depth values tend to have a wide range, with larger values being more common. This transformation reduces the impact of large depth variations and improves learning stability. Besides, the most important depth values for obstacle avoidance are the close ones. 
Although traditional segmentation architectures such as U-nets~\cite{unet} could have been an option, we were tied by the computational resources on the robot, which does not dispose of a dedicated computation module. Driven by this, we chose a small enough latent space of dimension 32, and the encoder consists of a 4-layer CNN with batch norms. We obtain a mean squared error at the end of the training on the training set of 0.011, and of 0.027 on the test set. We visualize the reconstructed depth using monocular input RGB image in Fig.~\ref{depth reconstruction}.

\subsection{Low-level locomotion control}

User-input velocities and collision-avoidance compensations are added into a commanded velocity $(v_x, v_\theta)$. The yaw velocity $v_\theta$ is converted to wheel velocity offsets using a differential drive model. The sagittal velocity $v_x$ is integrated into a reference trajectory for a model predictive controller over linearized wheeled-inverted-pendulum dynamics. The resulting optimal control problem is cast as a quadratic program~(QP) and solved with the open-source \textsc{ProxQP} solver~\cite{bambade2023proxqp}, using hot-starting for real-time performance. The QP solution provides wheel velocities that are forwarded to actuators, either in a Bullet~\cite{bullet} physics simulation or on the real robot.

\section{Experimental results}

\subsection{Obstacle avoidance in simulation}

\subsubsection{Pure-navigation environment}

We can assess the performance of our navigation environment by running episodes of maximum 20 seconds, with the joystick input fully forward, and logging the episode lengths. For a maximum episode length of 20 seconds, the visual navigation policy survived 15 seconds on average over 100 trials. To compare, we do the same experiment without correction, and obtain an average survival length of 4 seconds.

\subsubsection{Navigation with locomotion control}

To compare the performance of the navigation policy when controlling the locomotion controller between simulator and real world transfer, we found a place which has not been changed in the scene between the capture and the real world evaluation, as illustrated in Fig.~\ref{vignette}. We make it spawn in the same place as in the real world experiment, same orientation, and we prompt it to go fully forward. We evaluate if the agent successfully clears the obstacles. Note that there are two avoiding actions to take. First, it needs to turn to clear the wall corner, and then turn further to clear the black rack. 

\begin{figure}[t]
    \centering
    \includegraphics[width=\linewidth]{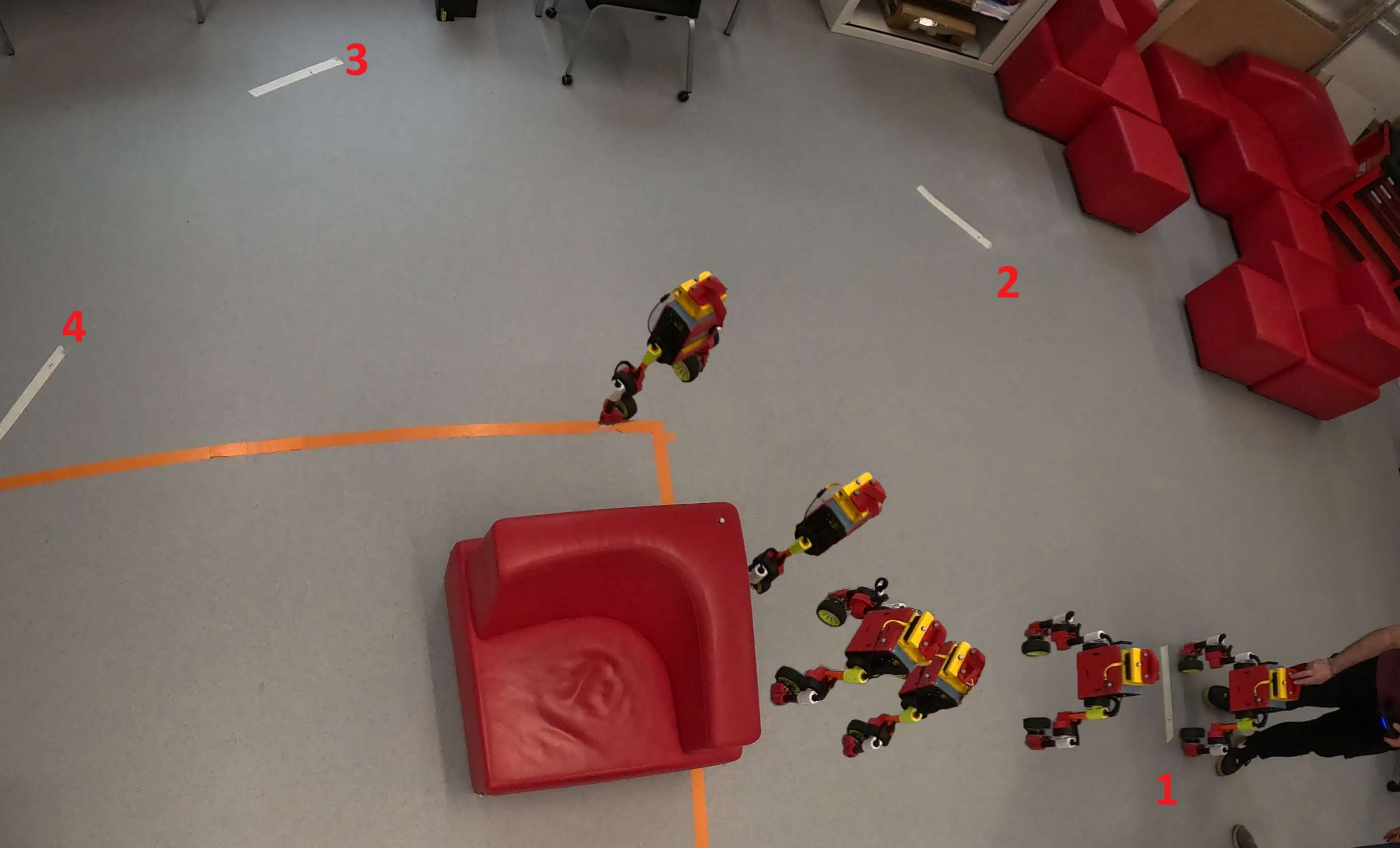}
    \caption{
        Illustration of the experimental setup, with starting positions 1 to 4 visible. During this trial, the robot stopped, went back, and turned towards a clear path while the input joystick command was a stationary forward throttle. The fifth marker is south in the figure.
    }
    \label{expe reelle}
\end{figure}

We obtain 100\% success rate of clearing the obstacles for both the pure navigation environment, and the navigation policy controlling the locomotion controller in simulation. We obtain 60\% success rate on the same experiment done on the real robot. The failure cases are when the navigation policy turns enough to not be able to see the obstacle in its field of view, but not enough to clear its body out of the way, suggesting an optimistic rate of turn. 

\label{sim2real section}
\subsection{Real-robot experiments}

We conduct experiments on an Upkie wheeled biped~\cite{upkie} equipped with a Luxonis OAK-D Lite camera. We only use the central 13~MP RGB fixed-focus camera, configured to output $640 \times 480$~px images. Those images are cropped to a central square and downsampled to the input resolution expected by the visual encoder.

\subsubsection{Training-room testing}

We first test obstacle avoidance in the robotics lab where the dataset was collected. Note that the furniture placement changed between scene capture and experiments, although obviously not the wall placement nor properties such as wall and floor colors. We positioned a red sofas (part of the training scene, but at a different location) at the center of the room, and marked five positions on the floor around it as illustrated in Fig.~\ref{expe reelle}. The user joystick input is set to full forward throttle from each position, so that the robot would collide with the centered sofa if no correction were applied. We do ten trials starting from each position, and evaluate the success rate at avoiding the obstacle.

We observed a 90\% success rate for the first position, 60\% for the second and the third one, and 40\% for the fourth one. Failure modes are similar to those described in Section~\ref{sim2real section}, where the robot clears the obstacle out of its field of view, yet not enough for his body to go through. Success rate for the fifth position was 0\%, with a yet-unseen failure mode in which the agent bluntly went forward with no significant correction, despite having the obstacle in sight at the beginning of the trial.

\subsubsection{Out-of-distribution testing}

To evaluate the robustness of the policy beyond the robotics lab where it was trained, we conducted tests in same-building office corridors and outdoor settings. These scenarios introduced new obstacles and different lighting conditions, allowing us to assess the agent's generalization capabilities.

In office corridors, the robot successfully avoided collisions, dynamically adjusting its trajectory to navigate around walls and obstacles. Notably, it managed to traverse narrow corridors while maintaining a forward heading, executing small corrective turns to stay on course. This behavior emerged despite the absence of such environments in the training data, demonstrating some degree of adaptability.

When tested outdoors, the robot encountered a wider variety of obstacles. It consistently refused to advance when facing large, immovable objects such as walls or parked cars, exhibiting cautious behavior even with obstacles absent from the training scene. However, it failed to react appropriately to certain smaller or visually complex obstacles, such as bicycles and grill gates, proceeding forward without correction. This suggests limitations in either its perception or learned decision-making when dealing with objects with fine structures or partially transparent elements.

\section{Conclusion}

We have trained a collision-avoiding policy that acts on RGB images and an \emph{implicit} environment model. The first step of our pipeline is an autoencoder-like architecture trained on a depth reconstruction task. The corresponding latent encodings are then forwarded to the policy itself, trained on an input-velocity compensation task, and its output is finally forwarded to a model-predictive locomotion controller. This approach transferred to a real wheeled-biped robot, with experiments showing repeatable collision-avoidance behaviors both in the training-set environment and in out-of-distribution environments.

Limitations and open questions to push the method forward revolve around generalization and continual learning. While we qualitatively witnessed a level of adaptability, with the robot traversing corridors despite having seen no corridor in its training set, the collision avoidance policy was not effective outdoors. The method displayed good repeatability, yet in both success and failure cases. A future iteration where failure cases could be improved with limited regressions on success cases would be significant.

\addtolength{\textheight}{-12cm}

\section*{Acknowledgement}

The authors wish to thank Justin Carpentier, Shizhe Chen and Antonin Raffin for valuable discussions, as well as Etienne Arlaud for valuable discussions and help with experiments. This work was supported by a grant overseen by the French National Research Agency (ANR) and France 2030 as part of the PR[AI]RIE-PSAI AI cluster (ANR-23-IACL-0008), and by the European Union through the AGIMUS project (GA no.101070165).

\bibliographystyle{IEEEtran}
\bibliography{references}

\end{document}